# Machine Learning Approach for Predicting Students' Academic Performance and Study Strategies based on their Motivation.

Fidelia A. Orji and Julita Vassileva

*Abstract*—This research aims to develop machine learning models for students' academic performance and study strategy prediction which could be generalized to all courses in higher education. Key learning attributes (intrinsic, extrinsic, autonomy, relatedness, competence, and self-esteem) essential for students' learning process were used in building the models. Determining the broad effect of these attributes on students' academic performance and study strategy is the center of our interest. To investigate this, we used Scikit-learn in python to build five machine learning models (Decision Tree, K-Nearest Neighbour, Random Forest, Linear/Logistic Regression, and Support Vector Machine) for both regression and classification tasks to perform our analysis. The models were trained, evaluated, and tested for accuracy using 924 university dentistry students' data collected by Chilean authors through quantitative research design. A comparative analysis of the models revealed that the tree-based models such as the random forest (with prediction accuracy of 94.9%) and decision tree show the best results compared to the linear, support vector, and k-nearest neighbours. The models built in this research can be used in predicting student performance and study strategy so that appropriate interventions could be implemented to improve student learning progress. Thus, incorporating strategies that could improve diverse student learning attributes in the design of online educational systems may increase the likelihood of students continuing with their learning tasks as required. Moreover, the results show that the attributes could be modelled together and used to adapt/personalize the learning process.

## I. Introduction

IN recent years, the use of online educational systems by both learning institutions and individuals for teaching and learning has become increasingly popular. Various courses in different programs and disciplines are made available to students through the systems. Despite the widespread use of the systems, issues of low engagement and high attrition rates are common among students learning with them. Thus, improving students learning experience in the systems through identifying their challenges and adapting appropriate learning interventions to address them is an important research field.

Student learning experience could be improved by identifying and understanding those at risk of failing or dropping out of university/course and providing timely interventions targetting their needs. To monitor student learning experience and identify those at risk, modelling their learning behaviours to predict whether they will succeed/fail in their studies is a well-known research area in student modelling. The modelling revealed that a variety of factors ranging from student motivation for learning, self-esteem, study strategies, basic psychological needs, and cognitive ability affect students' learning behaviours and progress. The goal of predictive modelling is to automatically detect these factors and respond with appropriate learning interventions to enhance learning. Most predictive models in the context of higher education seek to identify differences in students' learning in order to provide adaptive learning based on students' needs.

There is growing evidence that student motivation for learning affects their learning behaviour and performance. Moreover, research identified that student motivation changes during the learning process depending on the context [1]. The study explored a way of characterizing and detecting a particular student's motivation in order to manage and improve it dynamically. Thus, improving student motivation has attracted the attention of researchers as a tool for promoting student learning and progress. The impact of student motivation for learning is usually assessed based on dimensions of various motivation theories in education such as self-determination [2], self-efficacy [3], and self-esteem [4]. While there is an increased understanding of how each different motivational dimension affects learning individually, there is still a gap in understanding how they affect learning progress in combination. Thus, there is a need to investigate the combined influence of motivation variables (we will also call them "dimensions") on student study strategy and academic performance. This will guide eLearning systems designers in determining whether it is necessary to implement various strategies that target many motivation dimensions and it will also provide rich information for researchers demonstrating synergistic relationships among the key learning attributes.

To provide insight on how several motivation dimensions (intrinsic, extrinsic, autonomy, relatedness, competence, and self-esteem) predict learning performance and study strategy, we created and applied five supervised machine learning (ML) models. The algorithms used in building the models include Random Forest (RF), K-Nearest Neighbours (KNN), Decision Tree(DT), Linear/Logistic Regression (LR), and Support Vector Machine (SVM), as they are commonly used in solving regression and classification problems [5]. The models built were applied to a dataset of 924 university students from Chile, published in [6]. The best-performing regressor/classifier model for students' academic performance and study strategies was obtained using the test dataset. This study answers the following questions: Can intrinsic and extrinsic motivation, autonomy, relatedness, competence, study strategies (deep and surface), and self-esteem be utilized to identify at-risk students? How do intrinsic and extrinsic motivation, autonomy, relatedness, competence, and self-esteem influence students' study strategy? Among the five machine learning algorithms

F. A. Orji and J. Vassileva are with the Computer Sceience Department, University of Saskatchewan, 176 Thorvaldson Building, 110 Science Place, Saskatoon, SK S7N 5C9 CANADA (e-mail: fidelia.orji@usask.ca and jiv@cs.usask.ca).
.

investigated which regressor/classifier is the most successful at predicting student academic performance/study strategy? The results from this study revealed the impact of the aforementioned student attributes on study strategy and academic performance which can help in implementing appropriate interventions to enhance the students learning progress and retention rates. This also provides information for online educational systems designers about the dynamic role of incorporating strategies that could address various motivation dimensions in promoting students learning.

## II. RELATED WORK

### A. Student Motivation

One potential theory that is predominantly used in exploring student motivation for learning is the self-determination theory (SDT) by Ryan and Deci [2]. SDT recognizes the role of intrinsic and extrinsic motivations in facilitating people to achieve a specific goal. The authors postulate that meeting three psychological needs (*competence, autonomy, and relatedness*) of people promotes intrinsic motivation, self-regulation, and mental health, whereas when they are not met, they result in decreased motivation and well-being. The theory has been broadly applied in different learning contexts to explain how the need for growth and fulfillment of personal goals drives students' learning behaviours. According to the theory, certain antecedents such as the feeling of autonomy, competence, and relatedness/connection influence students' learning experiences which consequently affect their progress and performance. As such, the potential of these antecedents to effectively enhance student learning skills has been emphasized to show that when students learn different skills and gain mastery of tasks needed for specific goals, they would likely take actions that will lead them to achieve their goals. For instance, Feri et al. [7] applied multiple regression analysis to students' data collected using various validated questionnaires which include *autonomous* and *controlled/extrinsic motivation*. Students' academic performance was assessed using a 100-item multiple-choice test. The study reported that a 1% increase in students' autonomous motivation was linked to 15.2% improvement in students' academic performance. Chen et al. [8] revealed that addressing students' needs for autonomy, competence, and relatedness is likely to improve student engagement, course satisfaction, and performance in an online educational system. Thus, students' social context, skills for a specific task and their ability to direct their action to complete the task have direct influence on their learning performance. Moreover, several studies [9], [10] revealed that intrinsic motivation affects learning experience and academic performance of students. Students with high intrinsic motivation sustain for longer time their participation in academic activities to achieve required learning objectives more than those with low intrinsic motivation. On the other hand, extrinsically motivated individuals perform activities because of the external rewards that they will gain from completing the activities. The desire to perform an activity is driven by compulsion, rewards, or punishments rather than by pleasure and satisfaction obtained from accomplishing the activity. Extrinsic motivation is associated with high level of willpower and more engagement which could help at the initial stage of a task and as the learning process goes deeper may transform into intrinsic motivation for sustaining high-quality learning and creativity [11]. Thus, intrinsic and extrinsic motivation have a role to play in helping students to face learning challenges, understand processes, master, and apply skills learned in real circumstances.

### B. Predictive Modelling and Academic Performance

Academic performance has been described as fundamental for determining students' progress/success in their learning tasks. It is used to measure students' skills, their understanding of concepts, knowledge and their ability to accomplish set standard learning goals [12]. Academic performance is employed in assessing the extent to which learning institutions, educators, and students are meeting up their short and long-term learning goals. The assessment provides valuable information for educators that aids in decision-making, for example, in identifying weak students on the brink of failing and providing suitable counselling/allocating limited tutoring resources [13] or adapting appropriate learning interventions to assist students. Researchers monitor perceived relationships of academic performance with other learner attributes to detect factors affecting it and how students could be supported to improve their performance. Understanding the relationships also assists educators in improving their teaching.

Several studies have been conducted to explore the effect of some learner attributes on students' academic performance using various algorithms. For example, Chen et al. [14] used psychosocial factors, coursework grades and learning log data from advanced programming course in a higher education institution in predicting students' final grades. The researchers reported that coursework grades are the most significant factor followed by the total number of learning materials downloaded from the learning system. Similarly, a study in predicting students' academic performance using time on task, the total number of logins to a learning system, average assessment grades, and percentage of learning activities accessed revealed that the average assessment grade is the highest contributing variable, followed by the time on task [15]. The authors highlighted that the prediction model can be applied in identifying students that are likely to fail a course so that online educational systems can automatically initiate appropriate interventions which might involve both internal and external motivators. Jamjoom et al. [16] built models for predicting students' performance in an introductory computer programming course to detect at-risk students. The authors specified that the attributes used for the models were based on students' self-efficacy and they included cumulative high school grades, quizzes, midterm 1, midterm 2, practical evaluation, and final exam grades. The study results indicated that the decision tree model and support vector machine classifiers achieved the highest performance. The results also revealed that student self-efficacy is crucial for programming courses as it strongly correlated with practical evaluation grades. Marbouti et al. [5] created and compared seven prediction models to identify at-risk students. Scores of the following attributes; quizzes, attendance, weekly homework, team participation, project milestones, mathematical modelling activity tasks, and exams were used in the prediction. The Naïve

Bayes classifier and an Ensemble model produced the best results. Also, attributes such as race, family income, and university entry mode were employed in predicting student academic performance [17]. The study used Naïve Bayes, Ruled-Based and Decision Tree classification methods in discovering the best model for the prediction of academic performance based on the attributes. The result of the study revealed that race is the most influential variable, followed by family income and that the Rule-Based and Decision Tree models performed better than Naïve Bayes. Based on these reviewed studies on the prediction of academic performance, multiple factors affect students' academic performance.

Many examples can be found in the literature applying predictive modelling as a way of identifying at-risk students to provide interventions to assist them. For example, Greer et al. [18] applied predictive modelling built using high school GPA, actual course assessment grades, extrinsic versus intrinsic and deep versus surface study strategies to identify at-risk students and provided personalized learning supports and resources based on the students' needs. The researchers reported significant improvement in student performance. Also, Essa and Ayad [19] identified at-risk students using predictive models and segmentation and applied data visualization to gain diagnostic insights. The studies revealed that the use of predictive modelling combined with adaptive interventions can have significant impacts on student learning experience and performance. Current research trends in predictive modelling have shown that it can provide useful information about specific attributes or variables impacting the students' learning progress which can help online educational systems designers and educators in enhancing teaching and learning in higher education.

Although previous studies investigated several predictive models of student academic performance using various learning attributes, there is a need to explore the broad effect of autonomy, competence, relatedness, intrinsic motivation, extrinsic motivation, self-esteem, study strategies, and demographic attributes on student performance in higher education using supervised machine learning techniques. Moreover, few studies have investigated how the motivational constructs used in this research affect student study strategies. This study focuses on two widely recognized study strategies (deep and surface learning strategies) which have been shown to have conceptual and predictive advantages [18]. A few studies have shown that the motivational constructs and the study strategies used in this study have an impact on student performance. The predictors in our model were measured using established theories of self-determination, self-esteem and study strategies, which guarantees the generality of our model to various learning contexts.

III. METHOD

To determine the impact of some motivation dimensions and demographic attributes on the academic performance and study strategy of higher education students, we employed well-known machine learning techniques which are summarized below.

1. We performed some preprocessing to prepare the dataset we collected for analysis. Some data balancing techniques were applied.
2. We developed five supervised machine learning regressors and classifiers for predicting students' academic performance and study strategy.
3. We split our dataset into training and test sets. We trained and evaluated our regression models for academic performance prediction and classification models for study strategy prediction using 10-fold and 5-fold cross-validation respectively. We used the test sets to determine the performance of the models.
4. We compared the performance of the models built to determine the best-performing regressor/classifier.

*A. Data Description and Preprocessing*

Based on SDT theory, the quality and dynamics of people's behaviour are influenced by various forms and sources of motivation. These variations affect behavioural consequences such as persistence and performance experiences that accompany them. '*SDT therefore explicitly differentiates the concept in order to consider the varied effects of different types of motivation on such relevant outcomes*" [20]. Thus, different scales such as the academic motivation scale and satisfaction of psychological needs in education scale developed based on SDT theory are commonly used for assessing different sources of motivation and basic psychological needs in education.

The dataset for this research was acquired online [6]. It consists of 924 university students' data on intrinsic motivation, extrinsic motivation, and amotivation acquired using the academic motivation scale [21]; data on autonomy, relatedness and competence obtained through the satisfaction of psychological needs in education scale [22]; data on student self-esteem obtained using academic self-esteem scale [4]; deep and surface study strategies acquired using study process questionnaire [23]. The scales are self-scored and are the most frequently used instruments in education for assessing their various constructs because of their validity and reliability. The respective scale items that form intrinsic and extrinsic motivation, autonomy, relatedness, competence, self-esteem, deep and surface study strategies were combined for each of them to provide a general estimate of each of the features.

Various ways such as human observations, trained raters, and self-reports are used in providing ground truth labels for classification based on supervised machine learning. A study applied a self-report based on study process questionnaire to provide deep versus surface study strategy labels for a predictive model used in detecting at-risk students [18]. The study reported that deep versus surface study strategy, intrinsic and extrinsic motivation offer great predictive power for academic performance prediction. Based on previous studies which show that self-report can be utilized to provide ground truth labels for supervised classification [18], [24], we used the students reported study strategies to provide their study strategy labels for our classification algorithm.

Students' concurrent academic performance in the dataset was obtained from the administrative department of the university. Because motivation variables likely change over time, the concurrent academic performance seemed to be a more reliable metric than cumulative performance [21]. The dataset also contains demographic information of students which includes the year of study (ranging from 1 to 6), gender and age. The descriptive statistics of the dataset are shown in Table 1. Figs 1, 2, and 3 show the distribution of students based on their demographic information.

TABLE 1
DESCRIPTION STATISTICS OF THE DATASET

| Features | Mean | SD | Min | Max |
|---|---|---|---|---|
| Intrinsic motivation | 4.98 | 0.61 | 2.17 | 6.58 |
| Extrinsic motivation | 5.25 | 0.75 | 2.42 | 7.00 |
| Autonomy | 5.01 | 0.89 | 2.00 | 6.25 |
| Relatedness | 4.46 | 0.90 | 1.5 | 6.25 |
| Competence | 4.77 | 0.84 | 2.25 | 6.25 |
| Self-esteem | 4.17 | 0.17 | 1.75 | 7.00 |
| Deep Study strategies | 4.11 | 0.72 | 2.00 | 6.25 |
| Surface Study strategies | 3.32 | 0.79 | 1.50 | 6.25 |
| Study Year | 3.24 | 1.48 | 1 | 6 |
| Age | 22.83 | 3.36 | 18.00 | 44.00 |
| Academic Performance | 4.72 | 0.54 | 2.92 | 6.40 |

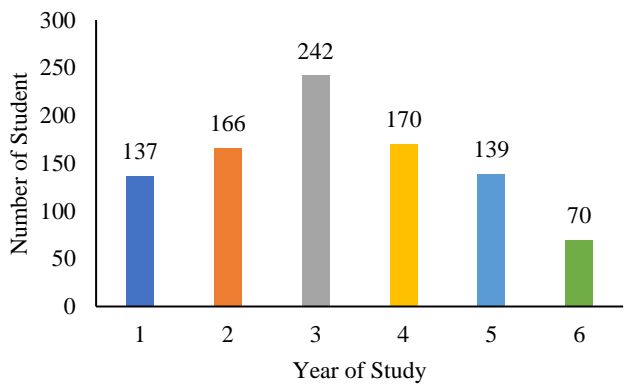

Fig. 1 Students distribution according to the year of study

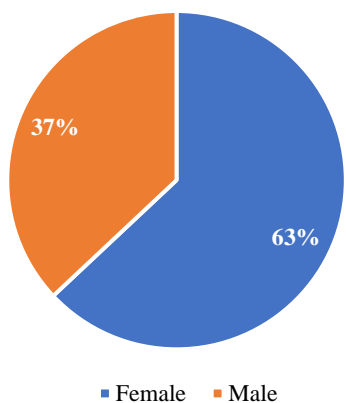

Fig. 2 Students distribution based on gender

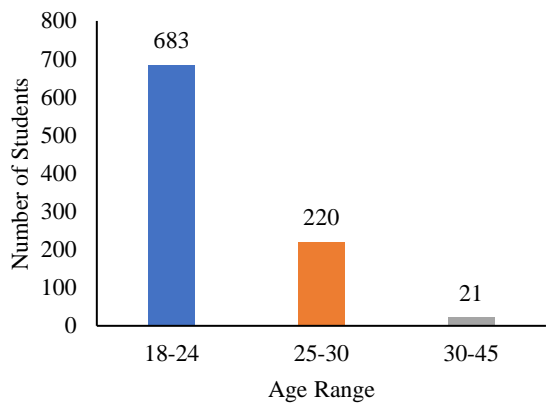

Fig. 3 Students distribution based on age

## IV. REGRESSION AND CLASSIFICATION EXPERIMENTS

Preprocessing of the dataset and prediction experiment in this study were performed using python and the scikit-learn library. We implemented and compared Random Forest (RF), Linear/Logistic Regression (LR), Support Vector Machine (SVM), Decision Tree (DT), and K-Nearest Neighbors (KNN), for both the regression and classification problems of student academic performance and study strategy. The dataset used in the model building was split into training and test sets in the ratio of 70%:30% for the regression experiment and 80%:20% for the classification. We examined many options in training and evaluating the regressors and classifiers using the training sets; the best results were obtained using 10-fold cross-validation for the regression and 5-fold for the classification. For the classification problem, out of 924 students' records in the dataset, 729 students were labelled as using deep study strategies, while 172 students used surface study strategies. This shows that our dataset for the prediction of students' study strategy was unbalanced. Because most machine learning algorithms need a balanced class distribution or an equal cost of misclassification, unbalanced data affects the learning process significantly [25]. The majority class is the deep study strategy while the minority is the surface study strategy. This means that the classification models could have overall high accuracy but only predict correctly the samples in the majority class while ignoring the minority class. To address this imbalance problem, we applied *random oversampling* techniques [26] using resample method in scikit-learn. The *random oversampling* method randomly duplicates some samples in the minority class depending on the number of samples the minority class needs to match the number of samples in the majority class. The datasets obtained after the sampling processes were used for prediction. Table 2 shows the size of data obtained using the technique.

TABLE 2
BALANCED DATASET FOR STUDY STRATEGIES PREDICTION

| Label | Original Dataset | Balanced Dataset (using oversampling) |
|---|---|---|
| Deep study strategy | 729 | 729 |
| Surface study strategy | 195 | 729 |

## V. RESULTS

Various evaluation metrics have been used in studies to understand the performance of regression models [6]. The performance metrics measure the closeness of predicted results to the actual values. Mean Absolute Error (MAE), Root Mean Squared Error (RMSE), and other metrics are frequently used in research investigations. In this research, we compared the accuracy of the regression models by computing their MAE. Using this measure, the average models' prediction errors can be directly interpreted (as the average difference between actual values and predicted values). The accuracy of the model is better when the MAE is lower. For the classification problem, we applied four frequently used evaluation metrics: accuracy, F1-score, precision, and recall [19]. The performance of both the regression and classification models was compared based on the above evaluation metrics using the test dataset.

The result of testing the five different machine learning models is shown in Tables 3 and 4. Table 3 shows the mean absolute error for each of the regression models. The decision tree model outperforms the other models while the KNN (the acronyms are defined in Table 3 and LR stands for Logistic Regression in Table 4) model produced the least accurate result. Table 4 shows accuracy across classifier models. Among the classifiers, RF achieved the best overall score in terms of accuracy, precision, recall and F1 followed by DT. The least accurate performance result was produced by LR. These results show that the features used in this research are sufficient to predict students' study strategies and academic performance.

TABLE 3
REGRESSORS' PERFORMANCE FOR ACADEMIC PERFORMANCE PREDICTION

| Regressors | Mean Absolute Error (MAE) |
|---|---|
| Random Forest (RF) | 0.3913 |
| Linear Regression (LR) | 0.4003 |
| Support Vector Machine (SVM) | 0.4026 |
| **Decision Tree (DT)** | **0.3777** |
| K-Nearest Neighbors (KNN) | 0.4242 |

TABLE 4
CLASSIFIERS' PERFORMANCE FOR STUDY STRATEGIES PREDICTION

| Metrics | RF | LR | SVM | DT | KNN |
|---|---|---|---|---|---|
| Accuracy | **0.949** | 0.589 | 0.599 | 0.880 | 0.685 |
| Precision | **0.949** | 0.569 | 0.632 | 0.892 | 0.686 |
| Recall | **0.950** | 0.570 | 0.611 | 0.885 | 0.687 |
| F1 | **0.949** | 0.568 | 0.587 | 0.880 | 0.685 |
| True Positive | 139 | 85 | 62 | 122 | 101 |
| False Positive | 18 | 69 | 92 | 32 | 53 |
| True Negative | 127 | 81 | 113 | 135 | 99 |
| False Negative | 8 | 57 | 25 | 3 | 39 |

### A. Predictive Features

To further examine the individual contribution of each of the features in predicting academic performance and study strategy, we applied RF (the best-performing model) to compute their feature importance. Computing linear regressions and examining the direction of the significant coefficient to determine the predictive power of the features were not used because the tree models (RF and DT) performed better than the linear models. As can be seen in Figs. 4 and 5, all the features are positive predictors of academic performance and study strategy. However, for academic performance prediction; study year, intrinsic and extrinsic motivation have higher predictive power than the other features, whereas for study strategy; relatedness, intrinsic and extrinsic motivation have higher predictive power. This means that addressing varied student needs through implementing design strategies which improve these features in educational systems will help to encourage students to progress better in their learning.

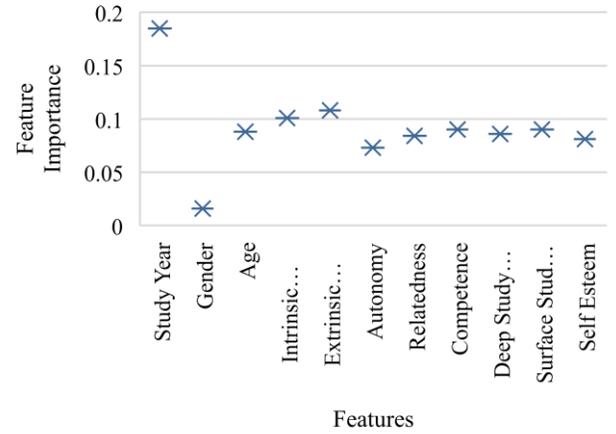

Fig. 4 Feature importance for academic performance prediction

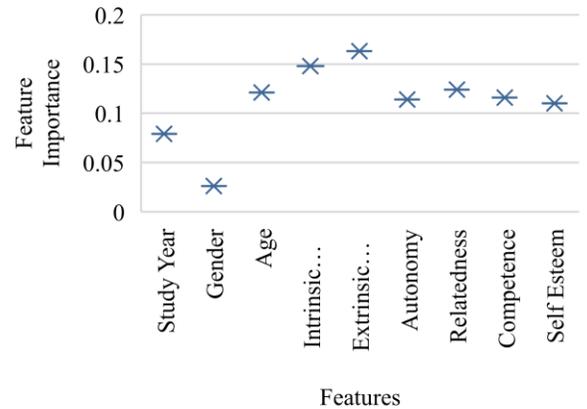

Fig. 5 Feature importance for study strategy prediction

## VI. DISCUSSION

According to education research, psychological factors such as motivation, basic psychological needs, self-esteem, study strategies, and demographic attributes, all influence student learning and consequently affect academic performance. Prior study applied the dataset used in this research to determine the motivational profiles of students and how the profiles associate with study strategies, basic psychological needs, vitality, and self-esteem [6]. Based on cluster analysis, the study identified four motivational profiles: *high intrinsic and extrinsic*, *high intrinsic and low extrinsic, low intrinsic and high extrinsic, and low intrinsic and extrinsic motivation groups*. The result of the profile analysis revealed that high intrinsic motivation groups showed higher scores for basic psychological needs, self-esteem, deep study strategy, vitality and academic performance while the low intrinsic motivation group reported high surface study strategy and low basic psychological needs regardless of their extrinsic motivation. The study established an association between identified motivational profiles and the learning attributes investigated.

The prior study identified an association between these learning attributes; student motivation (intrinsic and extrinsic), self-esteem, autonomy, relatedness, and competence [6]. This study took a different approach to examine the influence of these key learning attributes on student academic performance and study strategy using supervised machine learning approach. Using random forest, linear/logistic regression, support vector machine, decision tree and k-nearest neighbours, we demonstrated the feasibility of applying these learning attributes for predicting student academic performance and study strategy. To answer our research questions, the findings in this study revealed that these attributes are important to learning success as they provided good accuracy value in predicting student academic performance and study strategy. The best prediction model for detecting at-risk students based on academic performance is the Decision Tree followed by Random Forest. The best classification model for study strategy is the Random Forest followed by the Decision Tree; they both achieved much better accuracy values than the other models.

Furthermore, we found that all the learning attributes considered (as shown in Figures 3 and 4) positively contributed to learning success though at varied ranges. Intrinsic and extrinsic motivation are the most significant learning attribute for predicting academic performance and study strategy. On the other hand, the most predictive demographic feature for academic performance is the year of study while that of study strategy is age. Our result is in line with a previous study which revealed that intrinsic and extrinsic motivation offer great predictive power for student academic performance in a course [18].

Determining the predictive power of various learning attributes will help educators and online learning platforms in understanding attributes that need to be enhanced to improve students learning progress. The results of this research demonstrate effective contributions of intrinsic and extrinsic motivation, autonomy, relatedness, competence, self-esteem, study strategies, and demographic attributes toward academic performance. Based on the results incorporating different design strategies which include external motivators, discussion forums which allow students to connect and share knowledge with others (relatedness), autonomy (through allowing students to control their goals), competence (by providing relevant skills needed to achieve specific goals ), and fascinating instructional design that will attract students in the design of online educational systems will influence how committed students will engage in their learning resources and consequently their learning outcome. Moreover, the results also show that the attributes could be modelled together and used to adapt learning resources or interventions to enhance students learning. In this way, adequate and personalized support to individual students can be adapted to their changing needs. Varying the degree of support to each student depending on their specific learning context needs as well as their competency level is important in providing real-time support driven by artificial intelligence.

This research discovered interesting patterns on how demographic and psychological attributes interact to influence and affect study strategy and academic performance. A lot of factors affecting learning need to be understood for a better education to be provided. As such several researchers, utilized predictive techniques and tools to uncover hidden relationships among learning attributes. Selecting relevant and useful learner attributes for effective prediction of student academic performance has led to the development of various predictive models which were applied to interventions that helped to minimize students' failure rates [18], [19]. For instance, course-based predictive models have been developed using regression modelling techniques such as linear regression. Important variables such as the number of posts on discussion, the total number of quizzes completed, views of lessons, reports, current and previous grades, etc. were employed in building the models. The generability of the models can be hampered by the sample course variables utilized for model fitting [27]. The models will work well for courses that have relatively consistent structures. However, the models' prediction results may not supply any further information that could be interpreted by practitioners to design useful interventions, thus limiting potential benefits that institutions might derive from their data by developing predictive modelling for predicting student success. However, building models with generic features as shown in this study could lead to generation of models suitable for supporting the diverse needs of higher educational institutions and online educational systems while also allowing them to take full benefit of predictive analytics in applying interventions that will help to minimize students' failure rate.

*A. Limitation*

The models created in this research depend on data collected using self-report. We acknowledge that the use of self-report responses has strengths (in terms of validity and reliability) and biases. Our data on study strategies considered only deep and surface strategies, strategic learning was not considered. Although we have shown that the models built in this research can predict study strategy and academic performance with good accuracy using our available data, deep learning and ensemble approaches were not investigated to know if they will further improve the performance of the models.

VII. CONCLUSION

We applied the supervised machine learning (ML) approach to understand the predictive ability of some psychological attributes with respect to student academic performance and study strategy. Specifically, we implemented five ML regression and classification models and compared their performance to determine the collective impact of the attributes on study strategy and success. This study developed effective models that can predict student academic performance and study strategy using generic attributes, which means that the models can be applied across various courses in higher education or in predicting whether a student will graduate. In contrast, several existing course-specific prediction models may be hard to generalize to other courses or areas [16]. Based on the results of our models, motivational and other learner attributes provided good accuracy values in predicting study strategy and academic performance. The best-performing regressor in this study has MAE of 0.3777 in predicting academic performance while the best-performing classifier has F1 score of 94.9%. The results imply that addressing varied student needs by incorporating design strategies which improve the learning attributes investigated in this research into educational systems will facilitate students to make better progress in their learning. The models generated in this study can be applied by educational administrators in identifying students' study strategies and those at risk of dropping out of a course/higher education in order to provide necessary support and interventions.

Future eLearning systems designers should include appropriate motivational support through the use of various design strategies and machine learning models in the applications to provide automatic adaptive support for students.

ACKNOWLEDGMENT

F. A. Orji thanks the NSERC Vanier Canada Graduate Scholarship for funding my research.